# Partial train and isolate, mitigate backdoor attack

Yong Li ,Han Gao

May 2024

## 1 Abstract

Neural networks are widely known to be vulnerable to backdoor attacks, a method that poisons a portion of the training data to make the target model perform well on normal data sets, while outputting attacker-specified or random categories on the poisoned samples. Backdoor attacks are full of threats. Poisoned samples are becoming more and more similar to corresponding normal samples, and even the human eye cannot easily distinguish them. On the other hand, the accuracy of models carrying backdoors on normal samples is no different from that of clean models.In this article, by observing the characteristics of backdoor attacks, We provide a new model training method, partial training(PT), that freezes part of the model to train a model that can isolate suspicious samples. Then, on this basis, a clean model is fine-tune to resist backdoor attacks. We used three datasets and four backdoor attack methods to conduct our experiments. The experiments showed that PT performed very well in defending against backdoor attacks.

## 2 Introduction

In recent years, deep neural networks (DNN) have been widely used in many important real-world such as recognition[1], Computer Vision[2][3][4][5][6], and machine translation[7]. Nonetheless, DNNs have been shown to be vulnerable to potential threats at multiple stages of their life cycle. In reality, users often use data sets provided by third parties to train their models for some reasons. This also gives malicious parties the opportunity to deploy backdoors in the model by training it on data which provided by adversary. Intuitively, backdoor attacks aim to trick the model into learning strong correlations between trigger patterns and target labels by poisoning a small portion of the training data. Backdoor attacks can be notoriously dangerous for several reasons. First, backdoor data can infiltrate models in many situations, including training models on data collected from untrustworthy sources or downloading pre-trained models from untrusted parties. In addition, with the emergence of more complex and covert triggers [8][9][10], backdoor attacks have become increasingly difficult to detect. Poisoned samples are similar to their corresponding clean samples, and it is difficult for even the human eye to see the difference. Existing defense methods can be roughly divided into two categories based on samples: one is a method that requires additional clean samples, and uses clean samples to fine-tune[11], prune[12][13][14] or other operations[15][16] to eliminate or reduce the impact of backdoor attacks. The other is a method that does not require additional samples for defense, such as ABL[17], data augmentation[18][19] and distillation[20][15]. The effect of these methods is not ideal under certain attacks, because the backdoor attack methods are becoming more and more advanced cause some of the algorithms in the article are not guaranteed to complete its defence. Secondly, some algorithms are difficult to implement in small sample data. Some assumptions in these article may not be easy to meet in reality.

The author believes that the principle of backdoor attack is that the model changes its decision boundary to the effect that the attacker wants by learning the trigger. It is essentially over-fitting learning of the trigger, and the trigger is generally fixed or dynamic by following a certain mapping. Under this assumption, it is not difficult to find a shortcoming of backdoor attacks, that is, in backdoor attacks, this kind of over-fitting learning can be easily detected. Simply put, we can limit the performance of the model by freezing a part of the model, reducing the performance of the model on normal samples while maintaining the high accuracy of the model on backdoor samples. In this case, the gap between the loss functions of backdoor samples and normal samples will further increase, which also makes it easier for us to isolate suspicious samples to defend against backdoor attacks. Overall, the method in this paper has the following contributions:

First, this method does not require additional clean data sets to fine-tune or retrain a model.

Second, a new defense idea for backdoor attacks is proposed in this article. To the best of our knowledge, PT is the first defense method of this idea, which complements the existing defense methods.

Third, the method requires very little computing power and achieves state-of-the-art results.

## 3 Related Work

Trigger: Training a backdoored modules by adding triggers (patches)[21], one[22] or multiple pixel[23] to a part of the samples in the training data set, the model learns the knowledge of the backdoor trigger and changes its decision boundary. However, this form of trigger is not so covert, so more stealthier triggers were proposed in later backdoor attacks, such as the BLEND[24] method of injecting triggers through picture interpolation, then the Sinusoidal signal attack (SIG)[8], Reflection attack (Refool)[25] and Convex Polytope Attack[26] is proposed. As well as some attack methods that use items in reality as triggers[27]. Some recent attack methods have generated triggers that are difficult to detect by the human eye, such as WANET[9] and LIRA[10].

Poisoning methods: Poisoning methods can be roughly divided into two types depending on whether to change the ground truth of the poisoned sample. A method of changing the poisoning sample to the target class specified by the attacker to complete the backdoor attack, such as BadNet[21], we call this method dirty label attack. The other is to keep the label of the poisoned sample consistent with the original label, such as SIG[8] and others[28]. We call this method clean label attack. This method avoids the possibility of being detected to a certain extent. However, the corresponding success rate of poisoning is relatively small. It is necessary to optimize the generation of triggers or poison as many samples as possible to complete the attack while ensuring the concealment of the triggers.

Defense: Here we only classify defense methods according to whether they require additional samples. First, defense methods that require additional samples such as fine-tuning[11][12], retraining, pruning[13][14][12], clustering[29], or defense methods that use distillation that require additional samples[16]. The other type does not require additional samples and is defended through the characteristics of backdoor attacks, such as ABL[17] that isolates toxic samples based on the loss function, and combat backdoor through data augmentation like Strong data augmentation [19], Deepsweep [18], and novel way such as model connectivity repair (MCR) [30],Neural Attention Distillation [15]. There are also some defense methods based on recurrence triggers among them[31] [32] [33] [34] [35]

## 4 Method

In this chapter, we will introduce the overall method flow and the meaning of symbols and the entire defender process.

We assume that the data-restricted user used data provided by an untrusted third party that could lead to backdoor attacks. The attacker's goal is to generate a $\Delta$ and add it to sample x to make it a backdoor sample $x'$, so that the model trained on this data set contains a backdoor and output the triggered sample $x'$, to its target class $y_t$, $(x, y \oplus)$, or random class without compromising the accuracy of the model as much as possible, as shown in eq1. The adversary does not have the authority to specify the training model and model parameters but has the power to completely manipulate the data.

$$M'(g(x, \Delta)) = y_t \qquad (1)$$

The goal of the defender is to reduce the impact of possible backdoor attacks on the model and to increase model accuracy as they can. The defender has complete power over model selection and parameter manipulation, and the defender does not have any additional clean data sets for users to retrain and fine-tune.

Table 1: Summary of symbols

| Symbols | Explanation |
|---|---|
| $D$ | Suspicious third-party data sets |
| $D_s$ | Suspicious sample set after isolated |
| $D_c$ | Training set after isolation |
| $M'$ | Model used to determine whether a sample needs to be isolated |
| $M_s$ | Base model trained on suspicious third-party samples |
| $M_c$ | The final clean model |
| $CE$ | Cross entropy function |

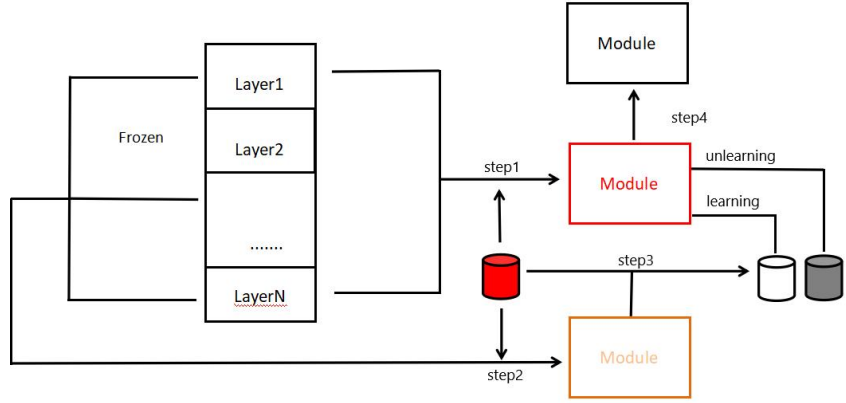

Figure 1: Stream Line:Defense process. The red cylinder is the data of the suspicious third party carrying the backdoor trigger. The white cylinder is the isolated data set, which may contain a small number of backdoor samples or clean data sets. The gray cylinder is the isolated suspicious sample data set. The red model is the baseline model, used to fine-tune a clean model. The yellow model is the discriminant model used to select isolated samples. Black is the final clean model.

Based on the above assumptions, we propose a new training method to resist backdoor attacks. This method can better isolate suspicious samples through the loss function by freezing a part of the model to train on the suspicious data set. The suspicious data set is then divided into two data sets according to the loss function, one of which contains a large number of backdoor attack samples, and the other data set contains a very small amount of poisoned samples or even a completely clean data set. On this basis, we train and unlearning on the two data sets respectively. We adopted four attack methods on three data sets, and the experimental results show that our defense method can reduce the ASR to about 0.001. The overall process of the method is shown in Figure 1. First, we train a model on the suspicious third-party data set D as the basic suspicious model $M_s$ through minimizing eq2, where CE stands for the cross entropy function and p is the probability of the corresponding category of the output.

$$Loss_1 = CE(p(x, y; \theta) | x, y \in D) \qquad (2)$$

In the second step, we initialize a model identical to $M_s$ and freeze the parameters of the first few layers, and train on D to obtain a model M'. In the third step, we calculate the loss function of each sample through the model M', and isolate a part of the suspicious samples based on their loss functions as $D_s$ for Unlearning, and the unisolated samples $D_c$ are used to maintain the model accuracy. The final training loss function is shown in eq3.

$$Loss_2 = CE(p(x, y; \theta) | x, y \in D_c) - CE(p(x, y; \theta) | x, y \in D_s) \qquad (3)$$

In the fourth step, we train and Unlearn the basic model M on the two data sets $D_s$ and $D_c$ respectively. Finally, we obtain a model $M_c$ with low ASR and high ACC. Through experiments, we found that the effect of separating third-party datasets by using model MS is not as good as model M' in most cases, which proves that PT can better assist in isolating suspicious samples. In the fourth step, in order to make the model more accurate, we use model MS as the basic model and conduct training on this basis.

## 5 Experiment

### 5.1 Dataset

(1) Cifar-10 [36] witch has 10 categories, each containing a total of 5000 samples ,totaling 50000 samples. We conducted two poisoning methods. 1 Using a non label clean approach, 250 samples were randomly selected from each category, with a total of 2500 samples accounting for 5% of the whole dataset, and placed in the target category called bird in this paper. 2 Using a label clean method, randomly select 2500 samples from the target label, accounting for 5% of the total dataset for poisoning.

(2) Mnist [37]. The MNIST dataset is from the National Institute of Standards and Technology (NIST) in the United States. The training set consists of handwritten numbers from 250 different individuals, of which 50% are high school students and 50% are staff from the Census Bureau. The test set also has the same proportion of handwritten digit data, but ensures that the author set of the test set and the training set do not intersect Two methods of poisoning were used. 1 Using a non label clean approach, 10% samples were randomly selected from each category and add them into the target category, called number 3 in this paper. 2 Using a label clean method, randomly select half of samples from the target label.

(3) Tsrd dataset. The TSRD includes 6164 traffic sign images containing 58 sign categories. The images are devided into two sub-database as training database and testing database. The training database includes 4170 images while the testing one contains 1994 images. All images are annotated the four corrdinates of the sign and the category(This work is supported by National Nature Science Foundation of China(NSFC) Grant 61271306). Due to the small number of samples included in some categories, we have expanded the sample size to five times the original size through data augmentation including centercrop, colorjitter, grayscale, hflip, vflip. We chose class 0 as target label and use the two methods mentioned earlier to poison the dataset.

| Attack | Dataset | Baseline ACC/ASR | ABL ACC/ASR | NAD ACC/ASR | SPT ACC/ASR | PT(ours) ACC/ASR |
|---|---|---|---|---|---|---|
| BadNet | minist | 0.98/1 | 0.65/0.043 | 0.974/0.004 | 0.98/0 | 0.965/0.007 |
| Blend | cifar-10 | 0.799/0.926 | 0.793/0.106 | 0.702/0.065 | 0.711/0.094 | 0.790/0.002 |
| SIG | cifar-10 | 0.795/0.976 | 0.672/0.011 | 0.672/0.078 | 0.810/0.100 | 0.792/0.002 |
| WANET | Tsrd | 0.720/0.971 | 0.667/0.103 | 0.361/0.047 | 0.695/0.036 | 0.732/0.001 |

Table 2

## 5.2 Attack Setting

We used three data sets: minister, cifar-10 and TSRD, where TSRD is a tinny sample data set. In minist, we use the BadNet method to conduct DLA attacks and select 10% of the data as poisoned data. We used WANET to poison 5% of the data set to perform DLA attacks on TSRD. On cifar-10, we carried out CLA attacks through the BLEND and SIG methods, selecting 1% and 3% of the samples for poisoning respectively. All training was conducted on Resnet-50. No data augmentation is used during the training process because we need to see the effect of the algorithm more intuitively. Each poisoning display is shown in Fig2.

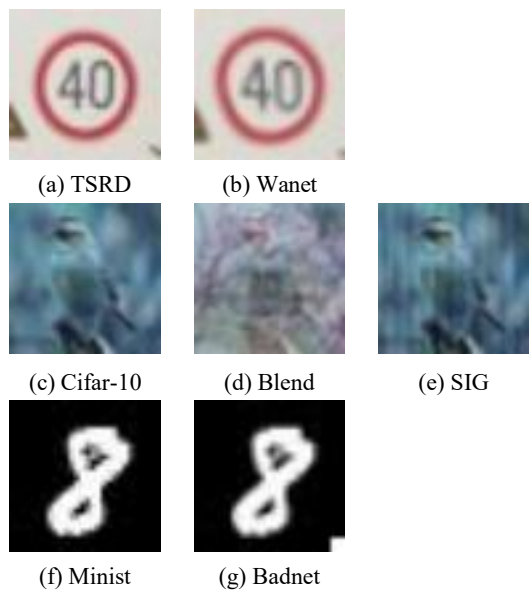

(a) TSRD    (b) Wanet

(c) Cifar-10    (d) Blend    (e) SIG

(f) Minist    (g) Badnet

Figure 2: a,c,f is the original picture of TSRD, CIFAR-10 and Minist respectively. b,d,e,g is its backdoor picture respectively, and the attack methods are Wanet, Blend, Sig and Badnet.

## 5.3 Defence Setting

We have selected several SOTA backdoor defense methods, including Super Fine-tuning(SFT), NAD and ABL. Among them, we select 20% of the corresponding toxic data set as an additional clean sample training set for SFT, and 10% as an additional clean sample training set for NAD. The corresponding parameter settings are set according to the recommended parameters in their article. For PT, We freeze the first 10% of the model layers for training to obtain the model M', and the isolation ratio is set to 1/10 of the training data. The training results are shown in TABLE2, and the ones with the best results have been highlighted in black.

### 5.4 Evaluation Metrics

In terms of measurement methods, we adopted Attack Success Ratio (ASR) and Accuracy (ACC). The first method measures how many of the test sets with triggers are judged as the target class. To avoid errors, we remove all samples of the corresponding target class in the test set when testing ASR. The second method gives the performance of the model on the normal data set, which is a traditional measurement method.

### 5.5 Result

In our defense approach, we set the sample isolation rate of suspicious datasets to 1/10. In the final training, for $D_c$, we set the learning rate to 0.001, while for $D_s$ our learning rate is 0.0001, and train for 10 epochs. We tested the performance of this defense method against four backdoor attacks. As shown in Table 2, we can see that PT has completed well defense in the four backdoor attack states. The ASR under the four attack methods has been reduced from nearly 100% to less than 1%. At the same time, the accuracy loss of the model does not exceed 3%.

## 6 Conclusion

This article provides a way to defend against backdoor attacks and achieve great results by isolating suspicious data and unlearning them by using partially frozen models. In the future we will try to conduct defense tests on more advanced backdoor attacks and introduce more better defense methods for comparison.